\documentclass[letterpaper, 10 pt, conference]{ieeeconf}  

\IEEEoverridecommandlockouts                              

\usepackage{graphicx} 
\usepackage{epsfig} 
\usepackage{mathptmx} 
\usepackage{times} 
\usepackage{amsmath} 
\usepackage{amssymb}  
\usepackage[linesnumbered,ruled]{algorithm2e}
\usepackage{caption}
\usepackage{subfigure}

\title{\LARGE \bf
Forest Tree Detection and Segmentation using High Resolution Airborne LiDAR
}

\author{Lloyd Windrim and Mitch Bryson$^{1}$
\thanks{This work has been submitted to the IEEE for possible publication}
\thanks{$^{1}$The authors are with the Australian Centre for Field Robotics, The University of Sydney, 2006, Australia. Correspondance:
        {\tt\small l.windrim@acfr.usyd.edu.au}}%
}

\begin{document}

\maketitle
\thispagestyle{empty}
\pagestyle{empty}

\begin{abstract}

This paper presents an autonomous approach to tree detection and segmentation in high resolution airborne LiDAR that utilises state-of-the-art region-based CNN and 3D-CNN deep learning algorithms. 
If the number of training examples for a site is low, it is shown to be beneficial to transfer a segmentation network learnt from a different site with more training data and fine-tune it.
The algorithm was validated using airborne laser scanning over two different commercial pine plantations.
The results show that the proposed approach performs favourably in comparison to other methods for tree detection and segmentation.

\end{abstract}

\section{INTRODUCTION}

Commercial forest growers rely on routine inventories of their forest in terms of the number of trees in a given area, their heights and other dimensions, often over large areas. With precise knowledge of the location of trees, their individual structures and the quantity and quality of wood they contain, resources can be utilised more efficiently during harvesting operations and supply-chain decisions can be planned more optimally. A combination of Airborne Laser Scanning (ALS) using manned aircraft \cite{Kaartinen2012,Ayrey2018} and Terrestrial Laser Scanning (TLS) using static, ground-based sensors \cite{Olofsson2016,Heinzel2017} is typically used to gather data for inventory, but these traditional techniques suffer from several limitations. Manned aircraft ALS typically results in point clouds with insufficient density to identify individual trees, TLS can only cover small areas of the forest. Recently developed UAV-borne LiDAR systems have demonstrated the ability to generate forest pointclouds with densities between ALS and TLS, and over large areas; issues still remain in how to extract inventory data (such as tree counts and tree maps) from these systems in an automatic way.

In this paper, we develop an automated approach to detecting, segmenting and counting trees in high resolution aerially acquired LiDAR pointclouds over plantation forests. Processing of lidar point clouds is an active research area in robotics and computer vision \cite{Li2010,Deuge2013,Weiss2011} where techniques must be robust to challenging, unstructured environments. This work draws from the robotics and computer vision literature to address the problems of detecting individual trees in a high resolution ALS pointcloud and segmenting each tree into its stem and foliage components. This segmented representation can be used to further derive a number of important attributes about each tree such as the crown height, stem diameter and volume of wood \cite{Pueschel2013,Liang2012}.

\begin{figure}[t]
\centering
\subfigure[Raw pointcloud. \label{fig:fp_1}]{\includegraphics[width=0.23\textwidth, clip=true,trim= 0 0 0 0]{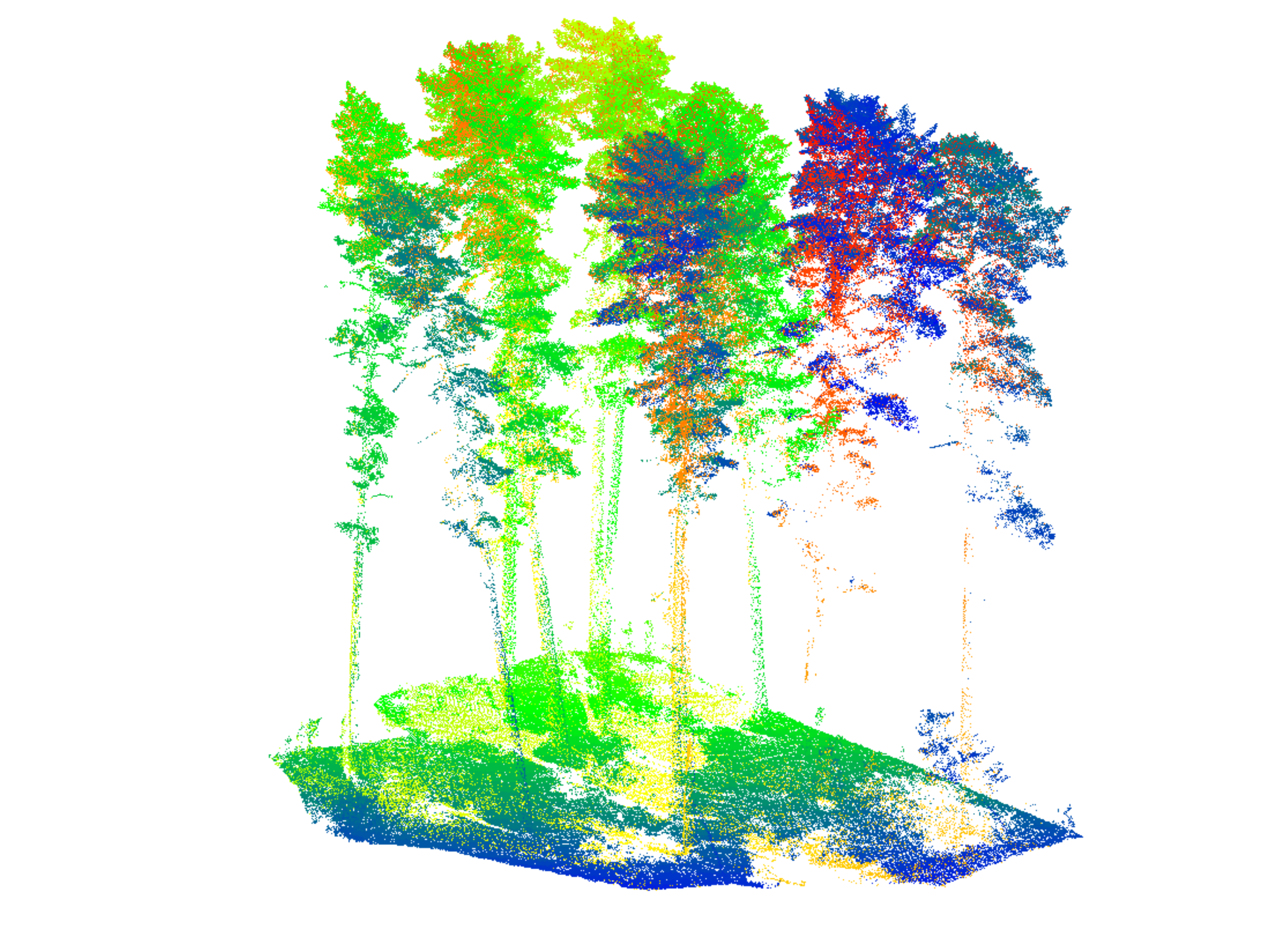}}
\subfigure[Identify ground points. \label{fig:fp_2}]{\includegraphics[width=0.23\textwidth, clip=true,trim= 0 0 0 0]{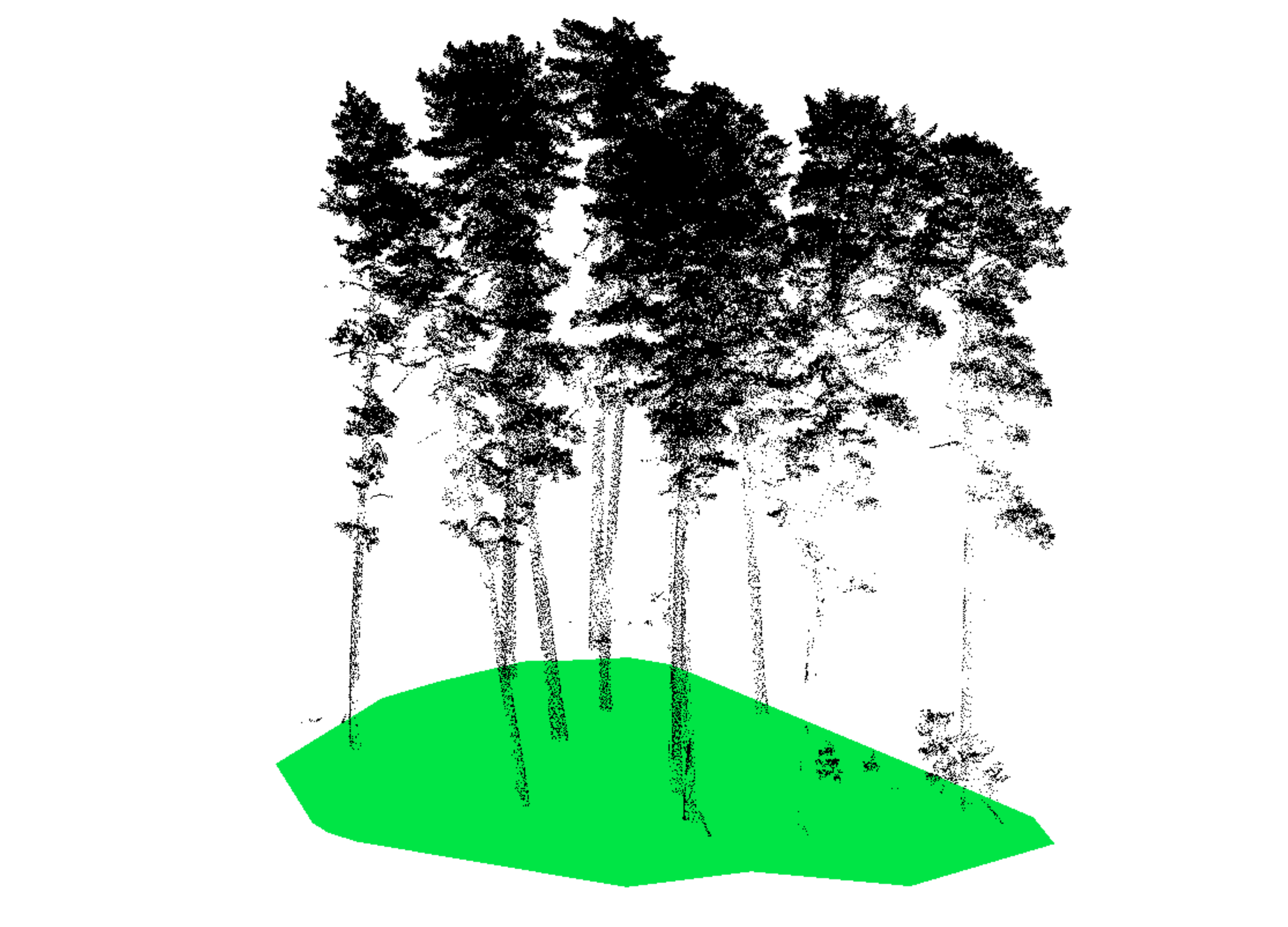}}
\subfigure[Detect trees. \label{fig:fp_3}]{\includegraphics[width=0.23\textwidth, clip=true,trim= 0 0 0 0]{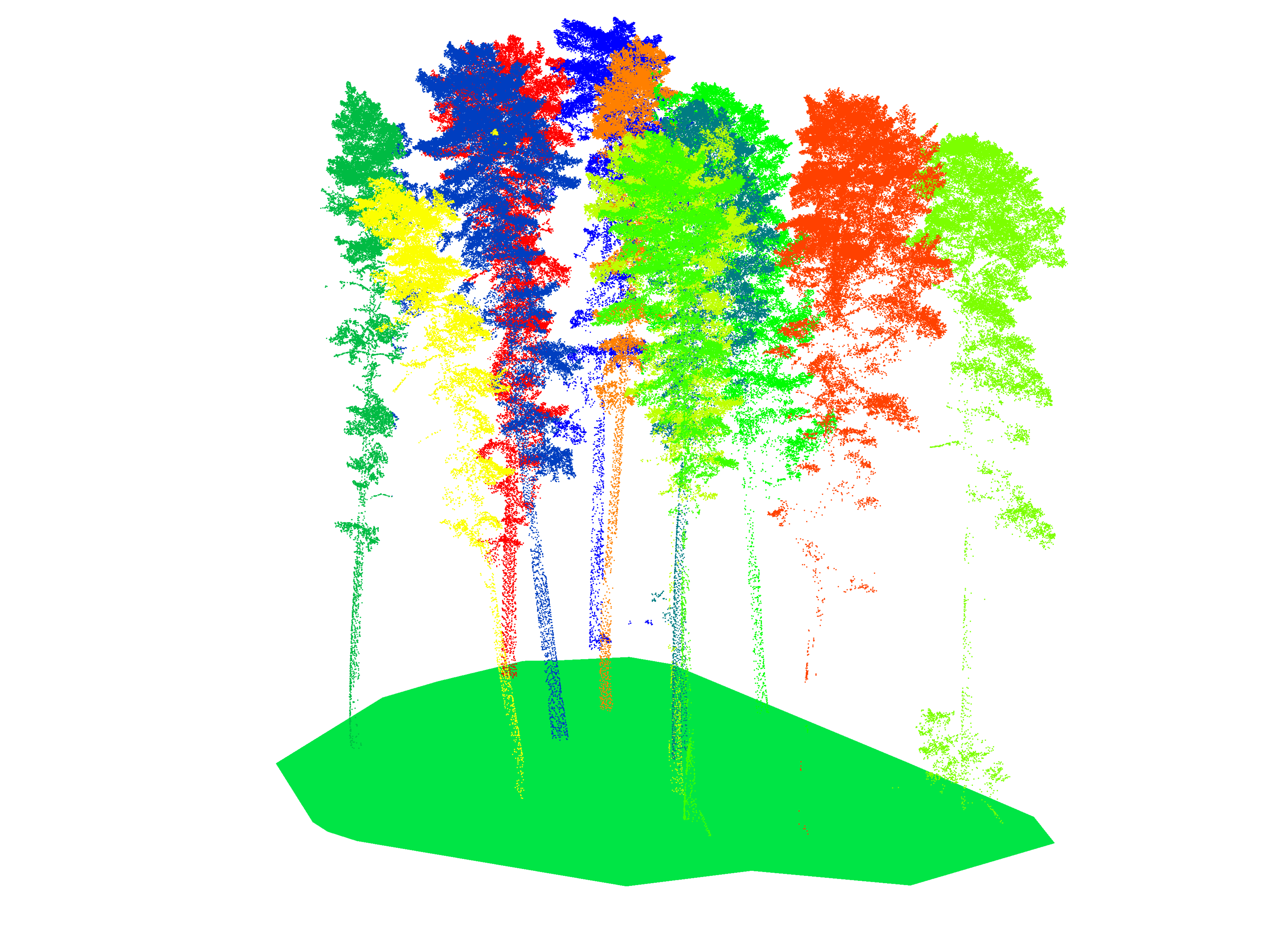}}
\subfigure[Segment trees. \label{fig:fp_4}]{\includegraphics[width=0.23\textwidth, clip=true,trim= 0 0 0 0]{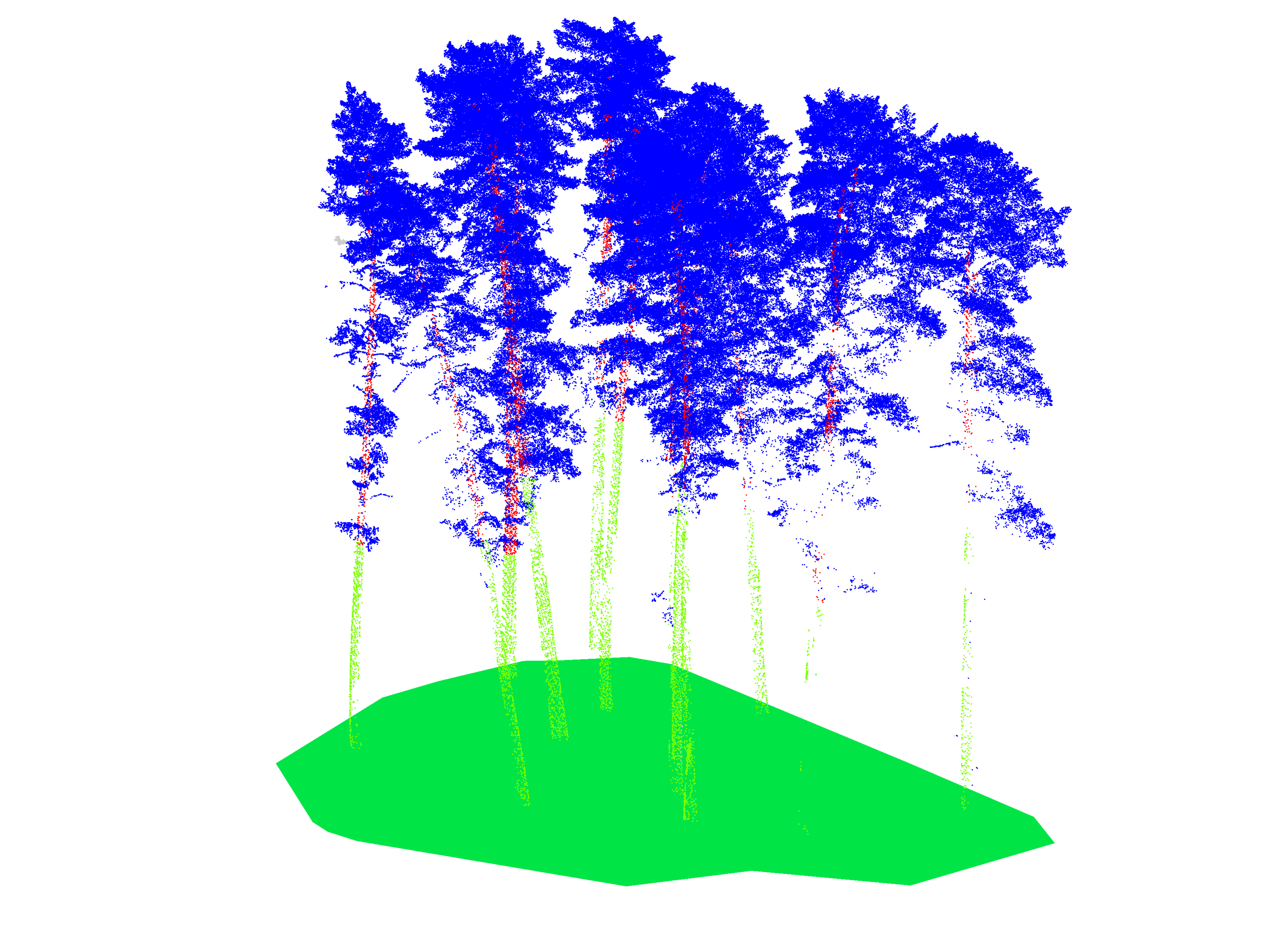}}
\caption{The different stages of autonomously processing a forest pointcloud to obtain an inventory.}
\label{fig:fp}
\end{figure}

The specific contributions of this work are: 
\begin{itemize}
\item{Detection of individual trees in a forest pointcloud.} 
\item{Segmentation of each tree into its components via per-point labelling of foliage, lower stem and upper stem.}
\item{An automated pipeline for 3D pointcloud processing for forest inventory which comprises the ground removal, detection and segmentation of each tree.}
\end{itemize}

Our processing methodology follows a machine learning paradigm based on state-of-the-art techniques in region-based convolutional neural networks (R-CNNs) and CNN-based 3D segmentation algorithms using a volumetric model of the forest derived from LiDAR pointclouds. Evaluation of our detection and segmentation algorithms is performed on high resolution ALS datasets acquired over two different commercial pine forests, with comparison against other methods for detection and segmentation of trees.

\section{PIPELINE FOR TREE DETECTION AND SEGMENTATION}

\begin{figure*}[t]
\centering
\includegraphics[width=0.95\textwidth]{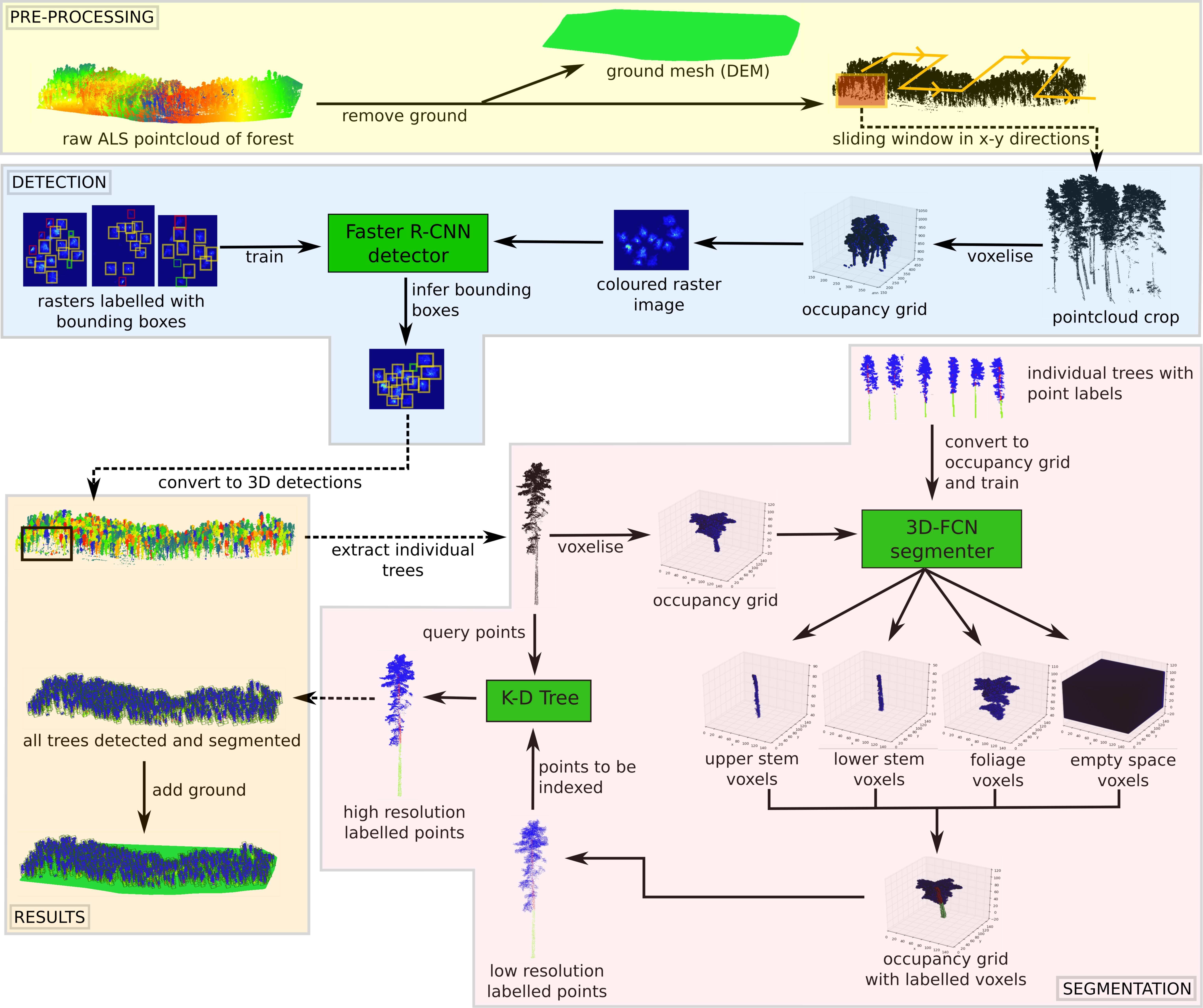}
\caption{Process for detection of individual trees in an ALS pointcloud and segmentation of their points into foliage, lower stem and upper stem.}
\label{fig:flowchart}
\end{figure*}

Given ALS data acquired over a forest, the aim of this pipeline is to detect the pointcloud subset associated with each tree in the forest, and predict a label for each point as either foliage, lower stem or upper stem. This process, summarised in Fig.~\ref{fig:flowchart}, involves removal of the ground points, object detection to detect cuboids that delineate individual trees and segmentation of the points those trees comprise into their semantic components using a 3D fully convolutional network (3D-FCN) designed to encode and decode occupancy grids.

\subsection{Ground Removal}

The ALS receives return pulses from the ground, as well as vegetation growing on the ground. It is useful to quantify the ground, for example, with a digital elevation model (DEM), to estimate forestry attributes such as canopy height. To simplify tree detection and segmentation, it is also important to remove all points associated with the ground and ground-based vegetation from the pointcloud. 

To estimate a DEM for the ground, the pointcloud is first discretised into bins in the $x$ and $y$ axes, and the point in each bin with the smallest height ($z$ axis) is stored. 
A regular $xy$ grid with four metre resolution that spans the point cloud is established. A K-D Tree \cite{Bentley1975} is used to find the closest four stored points to the centre of each grid cell, and the average height of these points weighted by their distance to the cell centre is calculated as the ground height for that $xy$ location. Once the height is computed for all cells in the grid, they are meshed using delaunay triangulation \cite{Delaunay1934} to output a smooth DEM. Finally, all points in the original pointcloud below a certain threshold above their $xy$ location on the DEM are removed, eliminating all ground points and most ground-based vegetation points.

\subsection{Detecting Individual Trees in a Pointcloud}

The structure in the data is leveraged to detect individual trees in the forest pointcloud. Trees are relatively uniform objects across a forest environment. They have vertical, cylindrical shapes with minimal overlap with adjacent trees and in most cases nothing can occlude a tree object in the vertical axis. Therefore, detections are made on 2D rasters from a bird's-eye perspective, using a CNN-based object detector designed for 2D imagery. The Faster-RCNN object detector \cite{Ren2017} is used to delineate trees by inferring bounding boxes in the $xy$-plane which are projected into three dimensional cuboids. 

To train the detector, 3D crops 
of land containing several trees are extracted from the forest pointcloud post removal of the ground points. These pointcloud crops are converted to 2D rasters which represent the vertical density of points at spatially discretised locations in the $xy$-plane. 
The vertical density is computed by summing the number of occupied vertical bins at each $xy$-location and dividing by the total number of bins for that location. 
Each 2D vertical density raster is mapped to a colour image, where tree objects have a distinctive appearance. Trees are annotated with bounding box labels. Two background classes are also labelled: shrubs and partial trees (i.e. those cut off when the plot was cropped out from the forest). These reduce the number of false positive detections. The coloured raster images and bounding box labels are used to train the Faster-RCNN object detector.

During inference, a window slides in the $xy$-plane and the corresponding cuboid (with the $z$-axis bounded by the maximum and minimum altitude of the data) is used to extract a crop of 3D points inside of it. The pointcloud crop is converted to a coloured image raster using the same process as for training. The trained Faster-RCNN model is used to detect bounding boxes around all trees, shrubs and partial trees in the raster. The window slides with an overlap so that there is a full tree for every partial tree detected. Tree class bounding box detections are accumulated, and redundant boxes that significantly overlap with others are discarded. The remaining 2D bounding boxes corresponding to the tree class are projected into 3D cuboids, and all 3D points within are identified as belonging to an individual tree.

\subsection{Segmenting Trees into Stem and Foliage}

Once pointclouds for individual trees have been detected, they are segmented into foliage, lower stem, upper stem or clutter components. A CNN is trained to segment the pointclouds in 3D, inferring one of these labels for each point. 
The architecture for the CNN is based on VoxNet \cite{Maturana2015}, which was a 3D-CNN designed for the classification of lidar scans of objects in urban environments, represented using occupancy grids. In this work it has been adapted for semantic segmentation, drawing from the structure of V-net \cite{Milletari2016}, which is a 3D fully convolutional encoder-decoder network for segmenting volumetric medical images represented as occupancy grids.

The 3D-FCN accepts a binary occupancy grid representing a single tree as input, with $150\times 150\times 100$ voxels of resolution $0.1\times 0.1\times 0.4$ meters in the $x, y$ and $z$ axes respectively. The network is trained to reconstruct four binary occupancy grids - one for each class (foliage, lower stem, upper stem and empty space - Fig.~\ref{fig:flowchart}). 
Every $xyz$ location is occupied in one and only one of the corresponding voxels across the four grids, with stem points having occupation priority over foliage points. If a location has no points in it then the voxel in the grid for the empty space class is occupied.

As in VoxNet, the first two layers of the network are 3D convolutional layers, with the first layer having 32 filters of size $5\times 5\times 5$ with a stride of two along all axes, such that the $150\times 150\times 100$ input is downsampled by half. The second layer has 32 filters of size $3\times 3\times 3$ with no downsampling. These two layers comprise the encoder, and the decoder comprises a mirrored version of these two layers with 3D deconvolutional layers instead. The second deconvolutional layer upsamples the data back to $150\times 150\times 100$. Each convolutional and deconvolutional layer precedes a leaky ReLU activation layer \cite{Maas2013}. There are skip connections between corresponding layers in the encoder and decoder to restore the resolution when upsampling. A final $5\times 5\times 5$ 3D convolutional layer maps the output of the decoder to the four target occupancy grids. 
A softmax nonlinearity is applied across corresponding voxels along the four occupancy grid outputs, treating them as one-hot vectors. A cross-entropy loss function is then used to compare predicted vectors to those in the target occupancy grids.

Pointclouds for individual trees are manually annotated by labelling points as either the foliage, lower stem, upper stem or clutter class (from a harvesting perspective the lower stem of a tree contains wood products of distinctive value from the upper stem). To train the 3D-FCN, each batch of single tree pointclouds are converted to binary occupancy grids for the input, and their labelled equivalent are converted to the four target binary occupancy grids. Points labelled as clutter comprise vegetation on the ground or foliage from adjacent trees. Clutter occupy voxels in the input grid are represented as 'empty space' in the target grid so that the network will learn not to reconstruct them. 
Each batch of tree pointclouds is converted to input and target occupancy grids on the fly so that the batch can be augmented with random rotations and flipping about the z-axis (which are done on the pointcloud prior to voxelisation).

During inference, 3D pointcloud crops from bounding box tree detections are converted to binary occupancy grids and passed through the trained 3D-FCN. The network outputs the four binary occupancy grids, one for each semantic component of the tree. The occupied voxels in the foliage, lower stem and upper stem grids are converted back to a single labelled pointcloud.

At this stage, the resolution of the labelled pointcloud is low because it was downsampled when it was converted to an occupancy grid. To restore its former resolution, the labels of the low resolution points are mapped to the high resolution points of the original pointcloud crop using a K-D Tree (Fig.~\ref{fig:flowchart}). Each point in the original crop queries the K-D Tree to find the nearest point in the low resolution, labelled pointcloud and inherits its label. If the distance to the nearest point exceeds a threshold, then the point is not given a label (these points are likely to be clutter). The result is a high resolution tree pointcloud with labels.

\section{EXPERIMENTAL SETUP}

High resolution LiDAR pointclouds were collected over commercial pine plantations in Tumut forest (October 2016) and Carabost forest (February 2018) in New South Wales, Australia using the Reigl VUX-1, a compact and lightweight scanner designed for UAV/drone operations. Data was collected at flying heights of 60-90m from the ground, resulting in pointclouds with a density of approximately 300-700 points per $m^2$. During experiments, the scanner was attached to a manned helicopter; future flights are expected to be performed using a commercial UAV.

From the Tumut site, 17 plot rasters comprising 188 trees in total were labelled with bounding boxes and 75 trees from across the site were also labelled at the point level. From the Carabost site, three plot rasters comprising 71 trees were labelled with bounding boxes and 25 trees were labelled at the point level. For testing, the Tumut and Carabost sites had three and one plot respectively where all trees had bounding box and point labels. The locations of the three test plots for the Tumut site were spread out across the forest and had 12, 8 and 11 trees, whilst the test plot for Carabost had 9 trees. To train the detectors for each test plot, 16 and 2 plot rasters comprising 176-180 and 62 trees were used for the Tumut and Carabost sites respectively. To train the segmentation networks for each test plot, 60 and 14 of the point-labelled trees were used for the Tumut and Carabost sites respectively. All labelling was done using open source software packages LabelImg \cite{Tzutalin2015} for bounding box labels and  CloudCompare \cite{Girardeau-Montaut2015} for point labels. 

\section{RESULTS}

To generate the 2D rasters for training the detector, a $600\times 600 \times 1000$ spatial grid with $0.2$m resolution was used, where the bins in the $z$-axis were accumulated before the raster was mapped to a colour image. Thus the input to the detector was a $600\times 600\times 3$ colour image. The Faster R-CNN detectors were trained for 10000 iterations through the data, using a learning rate of 0.003, momentum of 0.9, batch size of 1, Resnet-101 \cite{He2016} backend and Stochastic Gradient Descent (with momentum) optimisation.

The 3D-FCN segmentation network was trained until convergence (at least 3000 iterations). A learning rate of 0.001 was used for the first 500 iterations, and this was decayed to 0.0001 for the remaining iterations. The input shape of the data was $150\times 150\times 100$. The batch size and amount of data augmentation was limited by the GPU memory (11GB) and the number of training samples. For Tumut, the batch size was six with four additional augmentations per sample (30 in total per batch). For Carabost, the batch size was seven with three additional augmentations per sample (28 samples per batch). Training was done with the Adam optimiser \cite{Kingma2014} and classes were balanced in the loss function.

The detection component of the pipeline was compared against two ALS approaches for detecting trees. One found a canopy height model (CHM) for each test plot and then used marker-controlled watershed segmentation to detect individual trees \cite{Chen2006a}. The second technique used DBSCAN to cluster the pointcloud such that each cluster with more than a certain number of points was considered a tree \cite{Smits2012}. 

The segmentation component was compared against TLS methods used in mobile robotics applications. One method that used Eigen features coupled with a classifier \cite{Lalonde2006} was trained to label tree points as lower stem, upper stem and foliage. The other used a RANSAC approach to determine stem points \cite{Hogstrom1998}. Whilst the detection and segmentation components were evaluated separately, the same pointclouds detected as individual trees using the proposed approach were used as input for the segmentation experiments. Comparison methods for segmentation were given gold standard tree pointclouds as input.

Metrics used to evaluate detection were the precision, recall and F1 score for predicted tree pointclouds that had an intersection over union (IoU) with a ground truth tree pointcloud greater than 50\%. For the segmentation, the IoU was calculated separately for each class, as well as a combined stem class which treated upper and lower stem as the same (although models were still trained on upper and lower stem classes).

All processing was carried out on a 64-bit computer with an Intel Core i7-7700K Quad Core CPU @ 4.20GHz processor and Nvidia GeForce GTX 1080Ti graphics card. For the proposed method, on average, each segmentation model took four days to train and each detection model took 40 minutes to train. The average inference time for a single test plot was 50 seconds, with about 75\% of that time for the K-D Tree operations, which are dependant on the density of points.

\subsection{Tree Detection}

\begin{figure}[t]
\centering
\subfigure[Tumut Plot 1 pointcloud. \label{fig:det_result_1}]{\includegraphics[width=0.21\textwidth, clip=true,trim= 0 0 0 0]{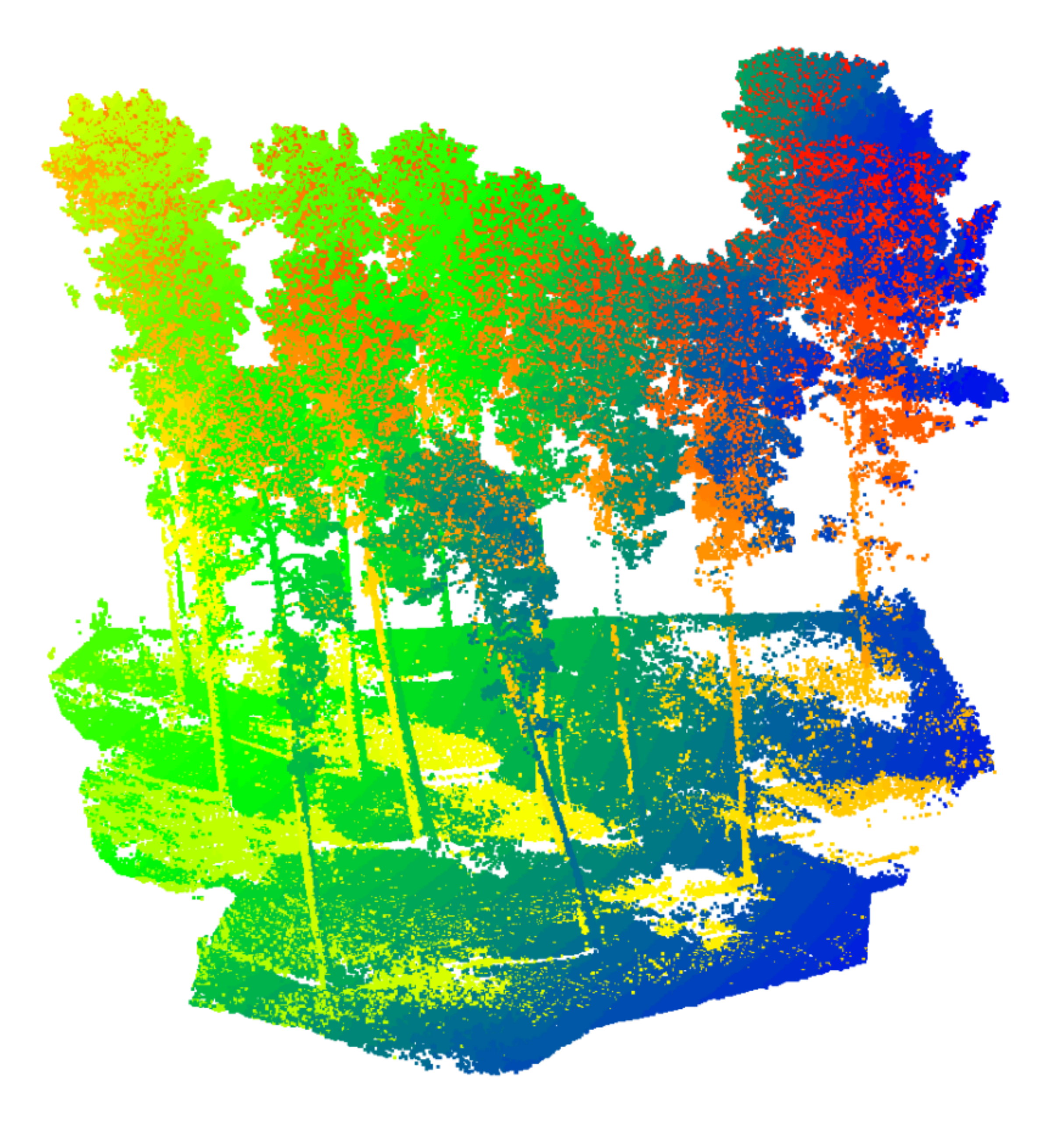}}
\subfigure[Tumut Plot 2 pointcloud. \label{fig:det_result_2}]{\includegraphics[width=0.21\textwidth, clip=true,trim= 0 0 0 0]{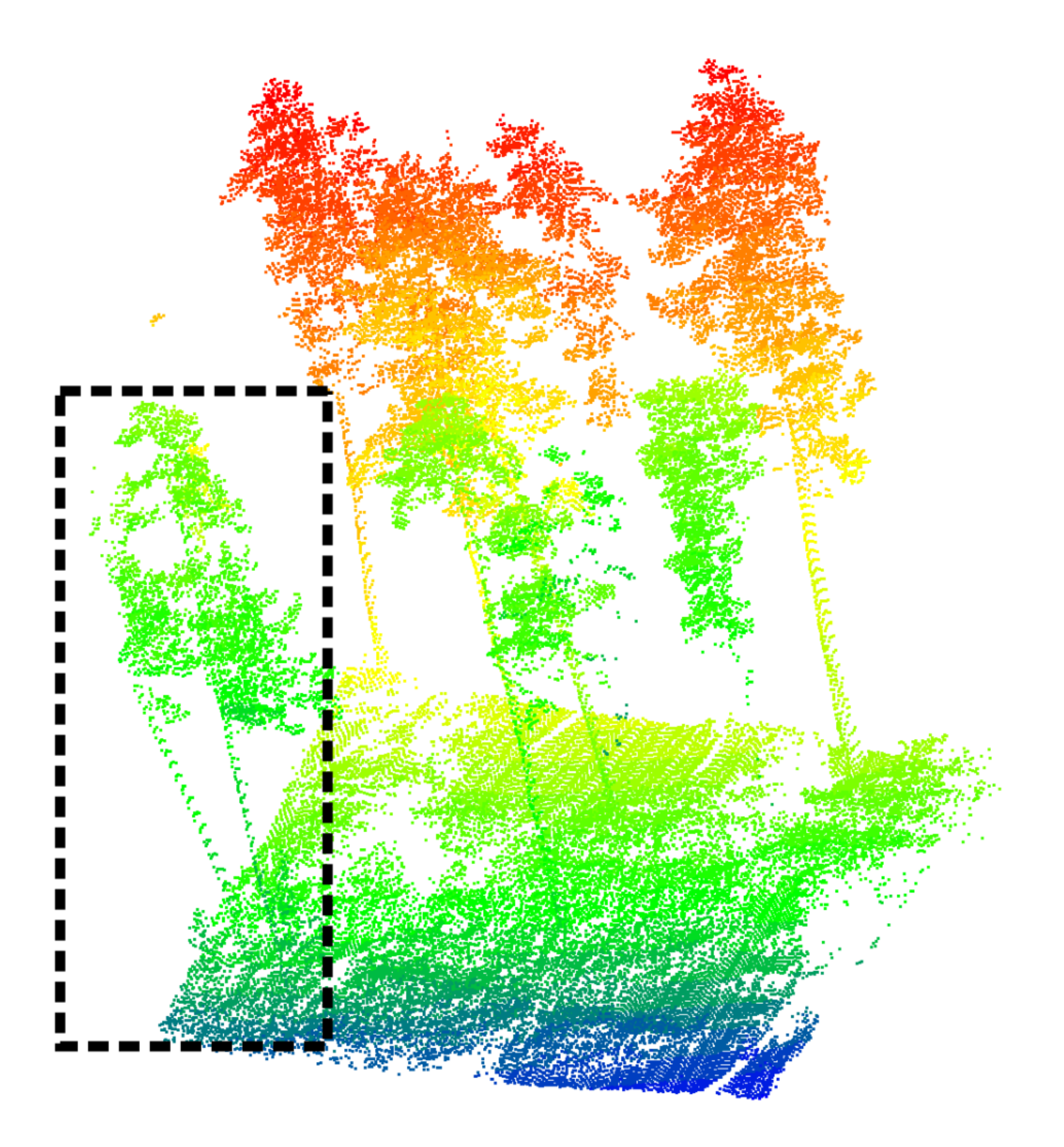}}
\subfigure[Tumut Plot 1 raster. \label{fig:det_result_3}]{\includegraphics[width=0.22\textwidth, clip=true,trim= 0 0 0 0]{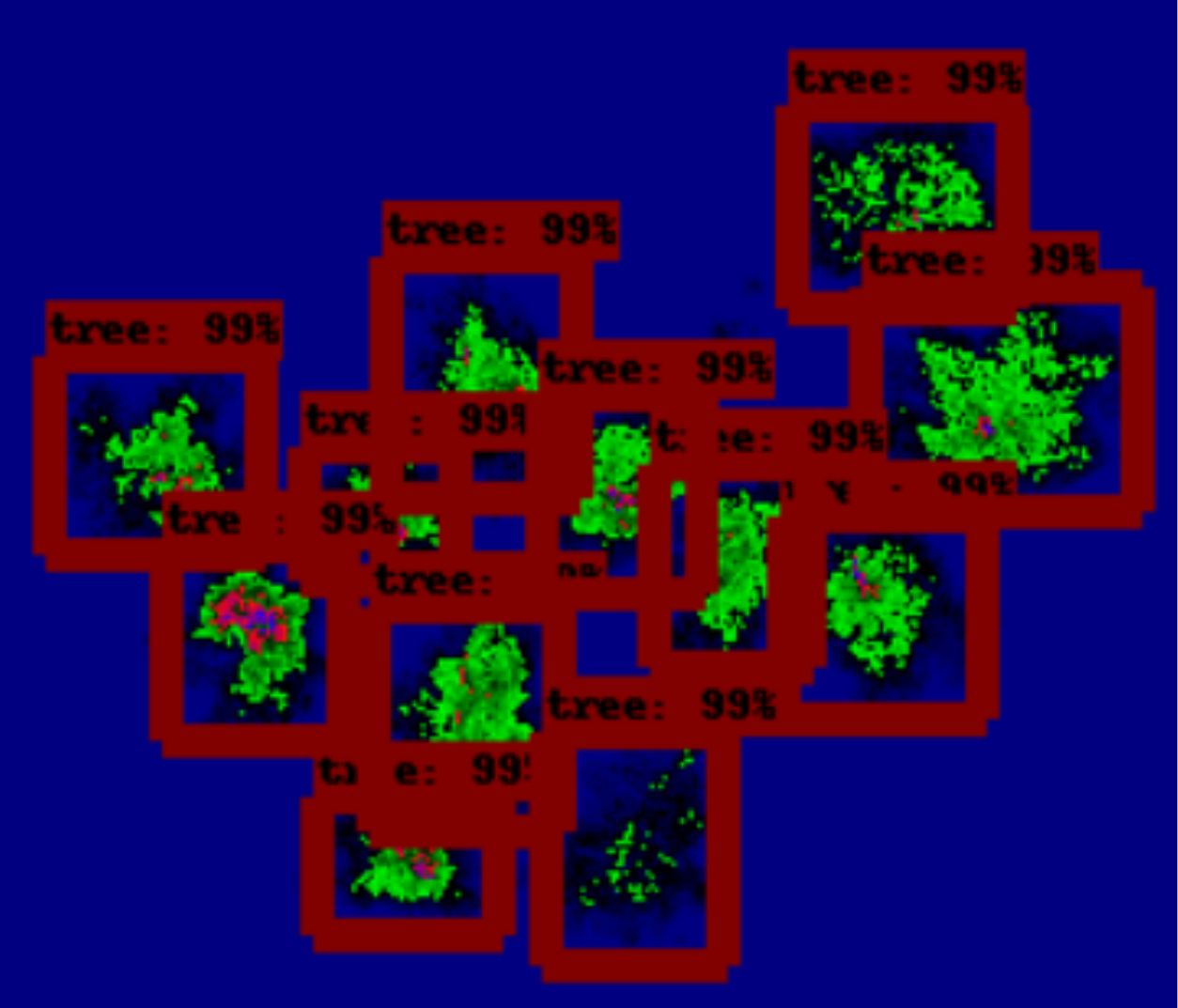}}
\subfigure[Tumut Plot 2 raster. \label{fig:det_result_4}]{\includegraphics[width=0.22\textwidth, clip=true,trim= 0 0 0 0]{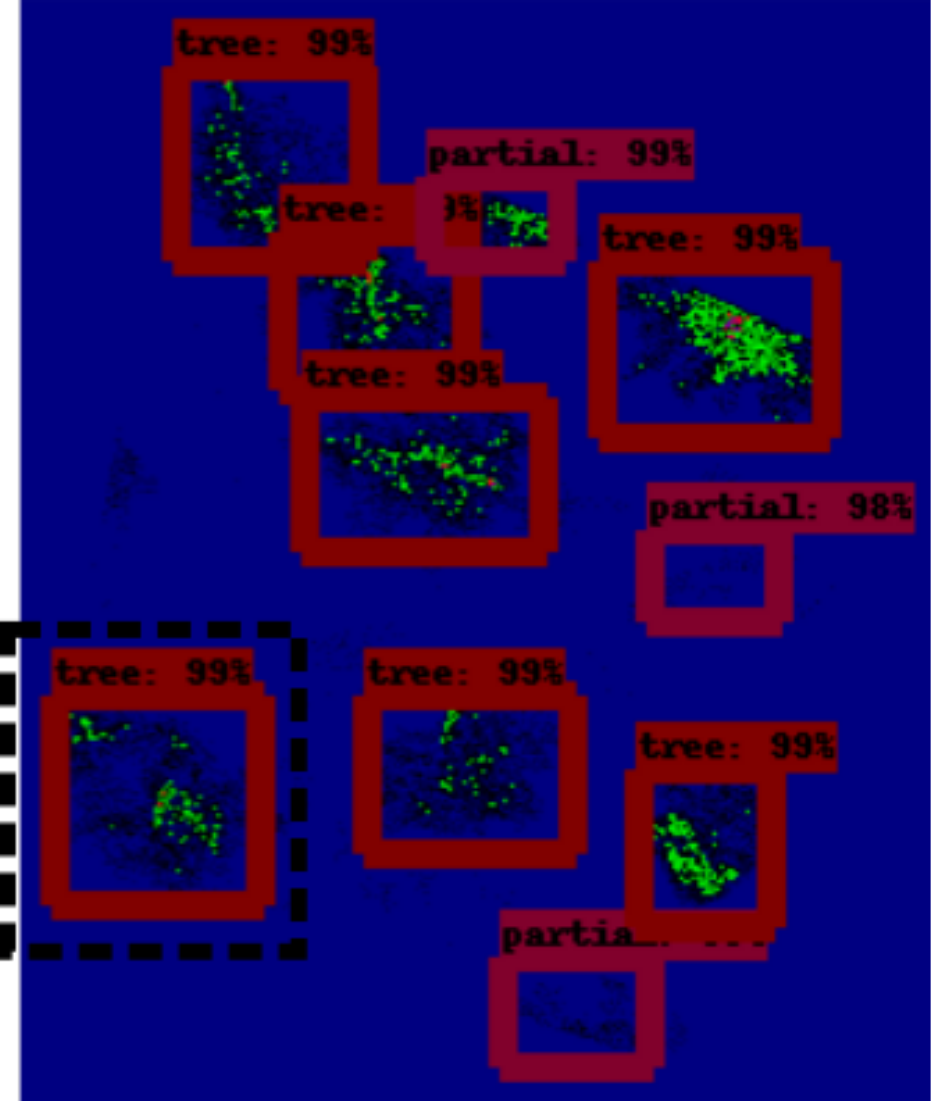}}
\subfigure[Tumut Plot 1 detections. \label{fig:det_result_5}]{\includegraphics[width=0.21\textwidth, clip=true,trim= 0 0 0 0]{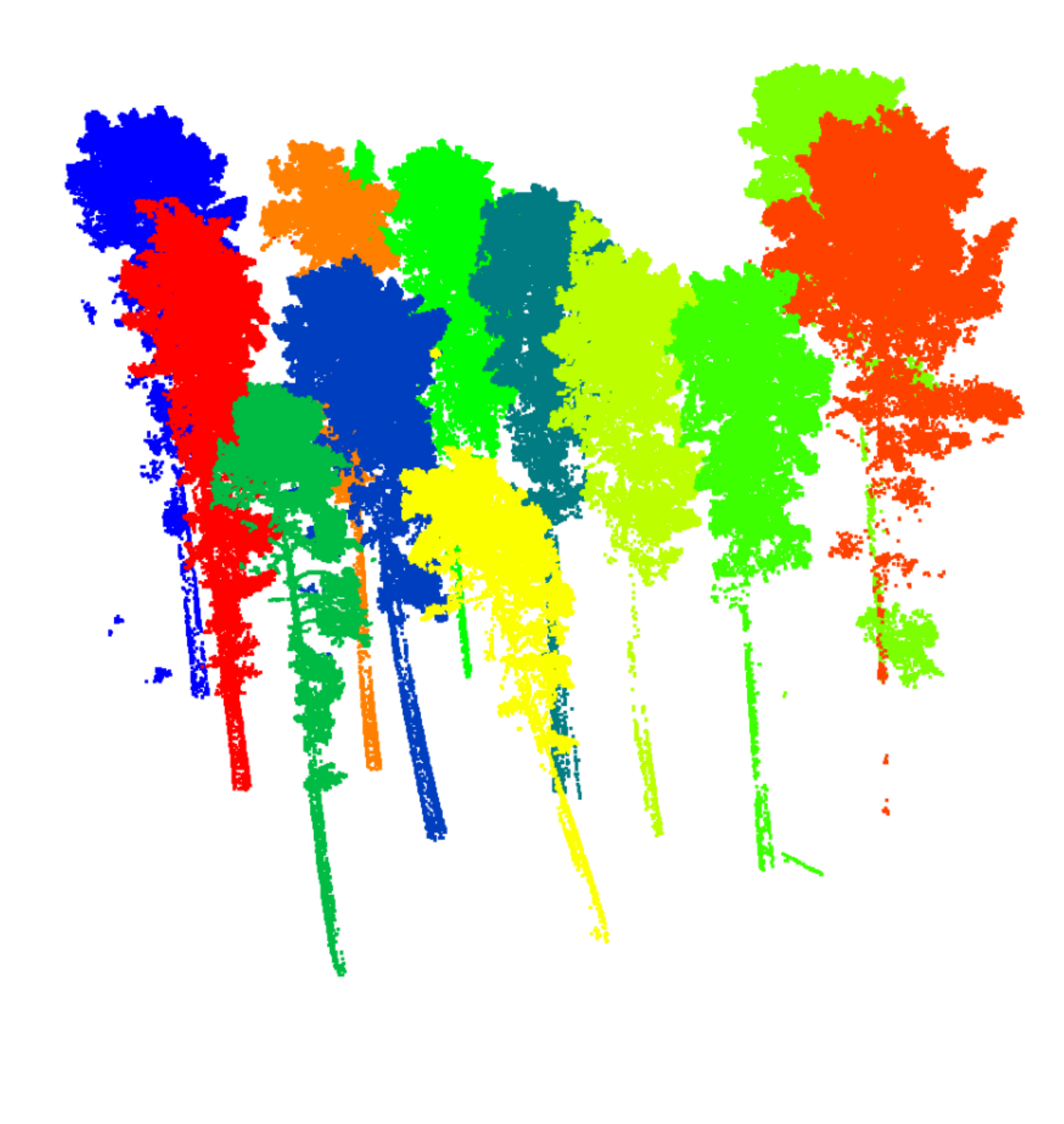}}
\subfigure[Tumut Plot 2 detections. \label{fig:det_result_6}]{\includegraphics[width=0.21\textwidth, clip=true,trim= 0 0 0 0]{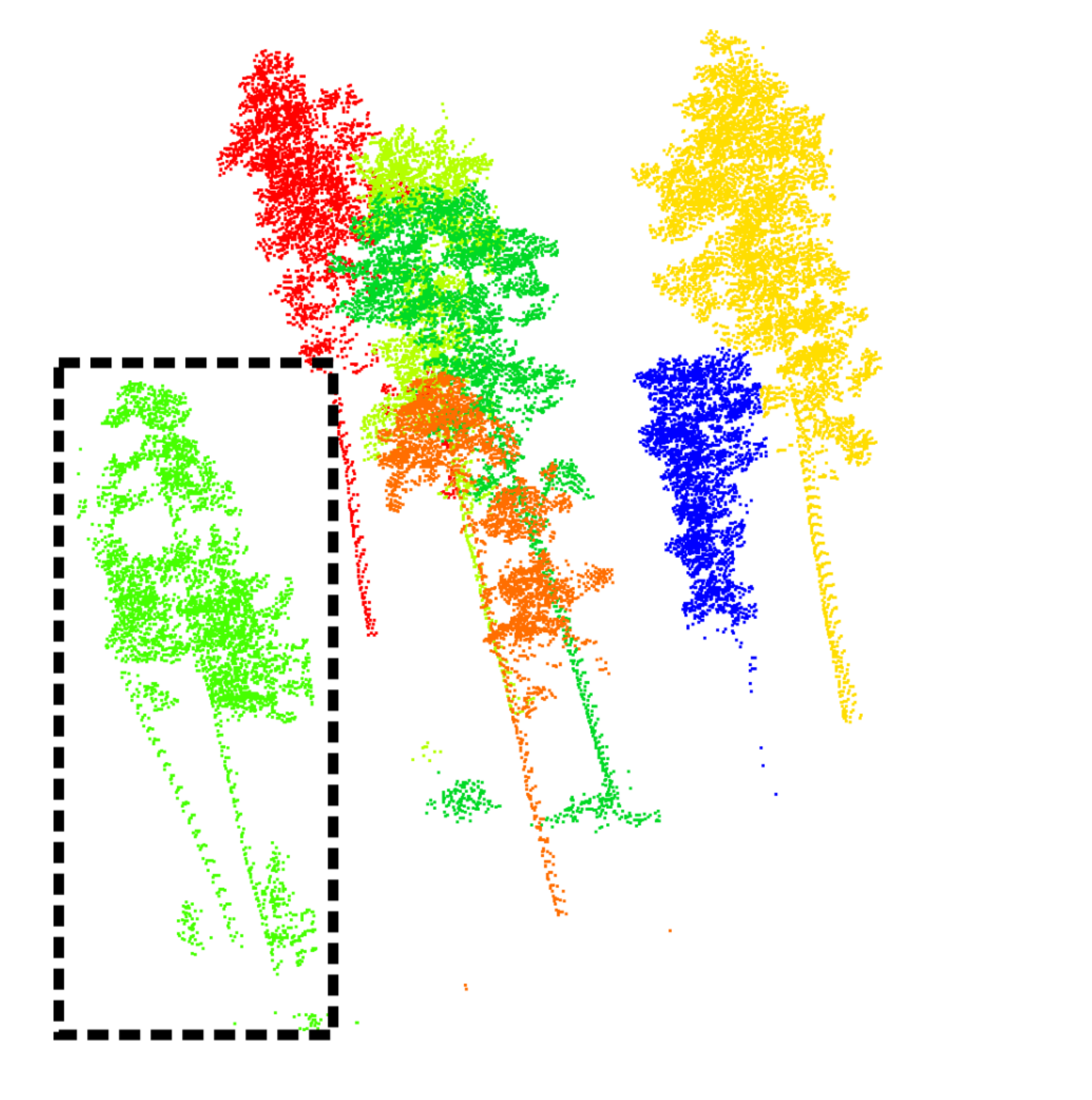}}
\caption{Visualisation of detection results for Tumut test plots 1 and 2. The dashed rectangle in plot 2 figures indicates an error where two trees have been detected as one.}
\label{fig:det_result}
\end{figure}

Table~\ref{tab:detection_results} and Figure~\ref{fig:det_result} show the individual tree detection results. The detection rates for the proposed method were high, with perfect scores for Tumut test plot 1 and Carabost. The proposed method only missed one tree from Tumut test plots 2 and 3, and it never predicted a tree where there was not one (the precision is 1.000 for all test plots).  For the Tumut sites, the CHM with watershed method had similar F1 scores to the DBSCAN approach, but achieved a perfect detection score on the Carabost data. The proposed method performed best overall.

\begin{table}
  \centering
  \caption{Detection results (precision, recall and F1 score) comparing proposed method with other ALS methods. The top performing method in each category is highlighted in bold.}
    \begin{tabular}{c|c|ccc}   
	\hline
	\textbf{Test Dataset} & \textbf{Method} & \textbf{Prec} & \textbf{Rec} & \textbf{F1} \\
	\hline
	Tumut Plot 1 & CHM + watershed \cite{Chen2006a} & 0.909 &  0.833 & 0.870 \\
	 (12 trees) & DBSCAN \cite{Smits2012} & 0.714 & 0.833 & 0.769    \\
	 & Proposed Method & $\mathbf{1.000}$ & $\mathbf{1.000}$ &  $\mathbf{1.000}$  \\
    \hline
    Tumut Plot 2 & CHM + watershed \cite{Chen2006a} & 0.556 & 0.625 & 0.588 \\
	(8 trees)  & DBSCAN \cite{Smits2012} & 0.750 & 0.375 & 0.500    \\
	& Proposed Method & $\mathbf{1.000}$ & $\mathbf{0.875}$ &  $\mathbf{0.933}$  \\
	\hline
	Tumut Plot 3 & CHM + watershed \cite{Chen2006a} & 0.727 & 0.727 & 0.727   \\
	(11 trees)  & DBSCAN \cite{Smits2012} & 0.643 & 0.818 & 0.720    \\
	& Proposed Method & $\mathbf{1.000}$ & $\mathbf{0.909}$ & $\mathbf{0.952}$   \\
	\hline
	Carabost & CHM + watershed \cite{Chen2006a} & $\mathbf{1.000}$ & $\mathbf{1.000}$ &  $\mathbf{1.000}$  \\
	(9 trees) & DBSCAN \cite{Smits2012} & 0.750 & 0.333 & 0.462  \\
	& Proposed Method & $\mathbf{1.000}$ & $\mathbf{1.000}$ & $\mathbf{1.000}$   \\
    \hline
    \end{tabular}%
  \label{tab:detection_results}%
\end{table}%

\subsection{Tree Segmentation}

\begin{figure*}[t]
\centering
\includegraphics[width=0.8\textwidth]{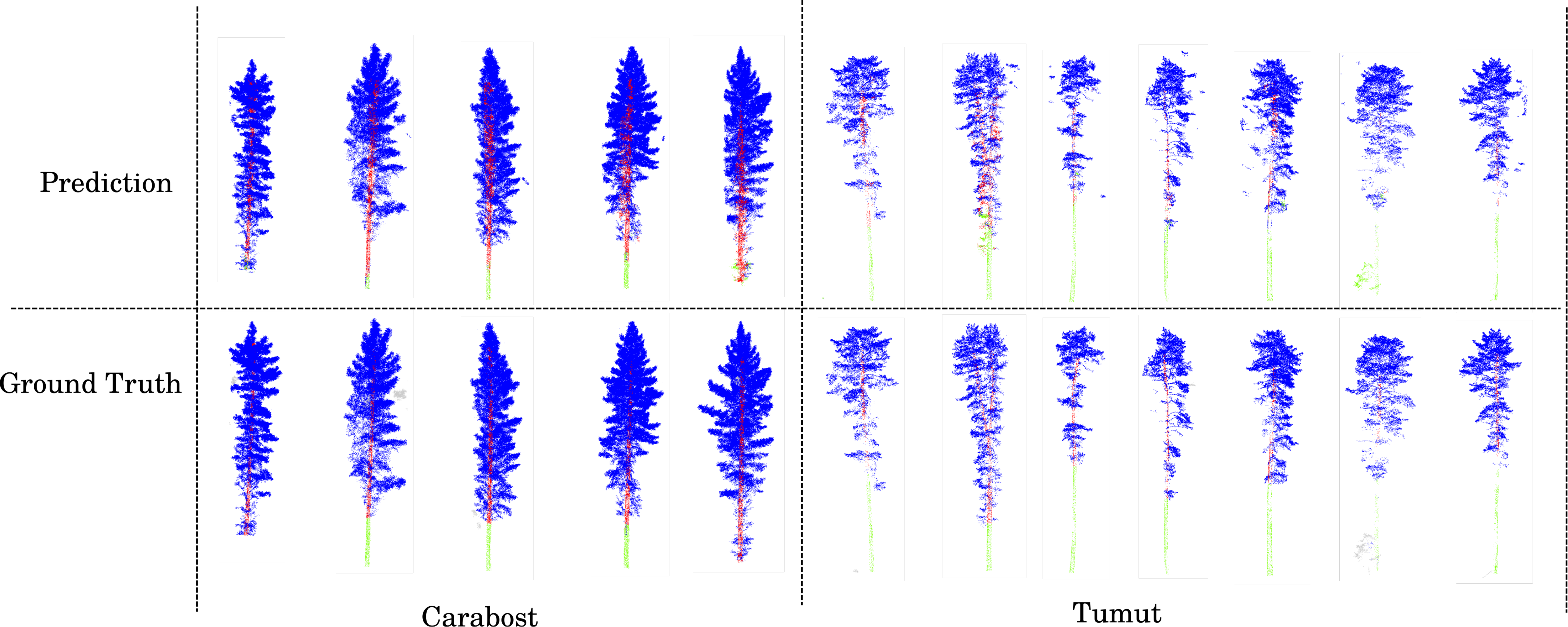}
\caption{Visualisation of selected tree segmentation results, with comparison to the ground truth.}
\label{fig:seg_result}
\end{figure*}

The segmentation results (Table~\ref{tab:seg_results} and Figure~\ref{fig:seg_result}) indicate that the proposed method performed best overall on the Tumut site data, particularly for the stem classes, where it outperformed the other methods by a large margin. The results for the Carabost site show that the RANSAC approach had the highest combined stem score. 

When a 3D-FCN model trained on Tumut data was used for inference on Carabost data without any fine-tuning, the overall result was a decrease in performance (Table~\ref{tab:seg_results_tf}). However, with fine-tuning of the network on the Carabost data, the results exceeded those where the model was trained solely on the Carabost data.

\begin{table*}[!t]
  \centering
  \caption{Segmentation results for the foliage, lower stem and upper stem classes, comparing the proposed method with TLS methods. The combined stem result is calculated by treating the lower stem and upper stem classes as a single class. The IoU is averaged over all trees in a given test plot. The top performing method in each category is highlighted in bold.}
    \begin{tabular}{c|c|c|c|c|c}
    \hline
	\textbf{Test Dataset} & \textbf{Method} & \textbf{Foliage}& \textbf{Lower Stem} & \textbf{Upper Stem} & \textbf{Combined Stem} \\
	\hline
	Tumut Plot 1 & Eigen features\cite{Lalonde2006} & $\mathbf{0.929\pm 0.043}$ & $0.050\pm 0.040$  & $0.025\pm 0.022$  &  $0.071\pm 0.055$    \\
	& RANSAC\cite{Hogstrom1998} & -  & - & - & $0.194\pm 0.086$   \\
	 & Proposed Method & $0.902\pm 0.057$ & $\mathbf{0.660\pm 0.240}$ &  $\mathbf{0.258\pm 0.134}$ &  $\mathbf{0.438\pm 0.169}$   \\
    \hline
    Tumut Plot 2 & Eigen features\cite{Lalonde2006} & $0.043 \pm 0.010$ & $0.053\pm 0.030$ & $0.003\pm 0.003$ & $0.087\pm 0.024$   \\
	& RANSAC\cite{Hogstrom1998} & - & - & - & $0.188\pm 0.100$ \\
	& Proposed Method & $\mathbf{0.873\pm 0.098}$ & $\mathbf{0.580\pm 0.337}$ & $\mathbf{0.260\pm 0.096}$ & $\mathbf{0.416\pm 0.141}$   \\
	\hline
	Tumut Plot 3 & Eigen features\cite{Lalonde2006} & $0.769\pm  0.253$ & $0.090\pm 0.091$ & $0.003\pm 0.004$ &  $0.104\pm 0.083$  \\
	& RANSAC\cite{Hogstrom1998} & - & - & - & $0.251\pm 0.174$  \\
	& Proposed Method & $\mathbf{0.924\pm 0.074}$ & $\mathbf{0.402\pm 0.253}$ & $\mathbf{0.3148\pm 0.158}$ & $\mathbf{0.407\pm 0.198}$   \\
	\hline
	Carabost & Eigen features\cite{Lalonde2006} & $\mathbf{0.957\pm 0.013}$ & $0.000\pm 0.000$ & $0.000\pm 0.000$ & $0.000\pm 0.000$  \\
	& RANSAC\cite{Hogstrom1998} & - & - & - & $\mathbf{0.427\pm 0.140}$  \\
	& Proposed Method & $0.847\pm 0.064$   & $\mathbf{0.218 \pm 0.270}$ & $\mathbf{0.289\pm 0.084}$  & $0.312\pm0.105$ \\
    \hline
    \end{tabular}%
  \label{tab:seg_results}%
\end{table*}%

\begin{table*}[!t]
  \centering
  \caption{Comparison of segmentation results for the Carabost site when transfering a model trained on the Tumut site (test plot 1). The IoU is averaged over all trees in a given test plot. The top performing method in each category is highlighted in bold.}
    \begin{tabular}{c|c|c|c|c}
    \hline
	\textbf{Method of training model} & \textbf{Foliage}& \textbf{Lower Stem} & \textbf{Upper Stem} & \textbf{Combined Stem} \\
	\hline
	Trained on Carabost & $0.847\pm 0.064$   & $\mathbf{0.218 \pm 0.270}$ & $0.289\pm 0.084$  & $0.312\pm0.105$    \\
	Trained on Tumut  & $0.742\pm 0.070$  & $0.111\pm 0.159$ & $0.130\pm 0.020$  & $0.147\pm 0.027$  \\
	Pre-trained on Tumut, fine-tuned on Carabost & $\mathbf{0.865\pm 0.065}$ & $0.136\pm0.170$ & $\mathbf{0.337\pm0.089}$ & $\mathbf{0.367\pm 0.114}$\\
    \hline

    \end{tabular}%
  \label{tab:seg_results_tf}%
\end{table*}%

\section{DISCUSSION}

The Faster-RCNN detector has many training examples of trees in the raster representation and can successfully generalise to unseen data. Even under many circumstances where the other two detection methods fail, such as if tree stems bend too much or fork in two, or if trees are very close to each other, the proposed detector can delineate individual trees. In Tumut test plot 2, the one tree that is missed by the proposed method is close to an adjacent tree, has a bent stem and the majority of its foliage distributed to the side of the adjacent tree (Figure~\ref{fig:det_result_2}). The detector incorrectly detects it as being attached to the adjacent tree (Figures~\ref{fig:det_result_4} and \ref{fig:det_result_6}). Similarly, in Tumut test plot 3, a tree with a small crown diameter that is close to an adjacent tree is misdetected as being a part of the adjacent tree. These are the only misdetections from the proposed approach. These incorrect detections have a trickle effect for the segmentation result. 

Regarding the segmentation results, there are more point-labelled trees for training from the Tumut site than the Carabost site, and hence the proposed method's segmentation result was better for Tumut than Carabost. However, the RANSAC method, which did not rely on training examples, performed better on the Carabost data than the Tumut data because the pointclouds have a higher density and the stems are more exposed. 

The lack of training examples for segmentation at Carabost was compensated for when the 3D-FCN network was pre-trained on the Tumut training examples and fine-tuned on the Carabost data. Whilst the foliage and combined stem results improved, the lower stem performance decreased. This is because more of the stem is denoted as lower stem in the Tumut site, and it is likely that the network assumes a similar structure for the trees in Carabost, labelling upper stem points as lower stem.

For the Tumut site, the upper stem was significantly harder to segment than the lower stem, which was reflected in the results for all methods. This is because it lies within the foliage, which blocks the LiDAR pulses, often resulting in large sections of stem missing from the scan. For the Carabost site, the structure of the trees is slightly different and the lower stem appears less frequently. The upper stem is also slightly more exposed than in the Tumut site. Thus the results of segmenting the upper stem were better than the lower stem. For both sites, the IoU scores for the stem classes are more sensitive to error than the foliage class because the stems occupy significantly less space. Slight misclassifications in the predictions cause large overlapping errors with the ground truth, resulting in big decreases in the IoU scores. 

The Eigen feature method was outperformed by the proposed approach on both sites. This was likely because it was designed for TLS data, where-as the ALS data has more noise and missing data due to pulses being occluded by thick canopy. The Eigen features are not as robust as the learnt 3D-FCN features. With sufficient training examples available from the Tumut site, the RANSAC approach \cite{Hogstrom1998} was also outperformed by the proposed method at this site. One of the major shortcomings of the RANSAC approach are that it does not work well when stems bend. It would also be negatively impacted by the missing sections of stem, which was more prominent at the Tumut site. 

One source of error in the proposed methods segmentation of the stem is due to the downscaling and upscaling of the pointcloud resolution. When the point-labelled trees are converted to low resolution occupancy grids, the stem classes have priority over foliage. The network is trained on this data and when the low resolution pointcloud is mapped back to the high resolution pointcloud, the stem points cover a wider space and encroach on the foliage class (see the Carabost trees in Figure~\ref{fig:seg_result}). Whilst the recall for stem points remains high, the precision gets negatively affected.

\section{CONCLUSION}

This paper presented a method for detecting and segmenting trees in high resolution airborne LiDAR. Using Faster-RCNN, trees were detected in a 2D coloured raster representation from a bird's-eye perspective. Once detected, trees were segmented in 3D at the pont-level into their stem and foliage components using a 3D-FCN and KD-Tree. Overall, the proposed approach outperformed other methods for tree detection and segmentation. It was also shown that pre-training a 3D-FCN on data from a site with more training examples can improve results. 

Future work will consider real time algorithms for deploying tree detection and segmentation on a UAV. Such a capability would enable aerial robotic applications in forestry such as targeted tree inspections, delivery of pesticides to the upper tree canopy and aerial robotic tree pruning.

\addtolength{\textheight}{-12cm}   




\section*{ACKNOWLEDGMENT}
This work was supported in part by Forest and Wood Products Australia research grant PNC377-1516. Thanks to David Herries, Susana Gonzales, Christine Stone and Interpine New Zealand for providing access to airborne laser scanning datasets.



\bibliographystyle{IEEEtran}
\bibliography{ICRA2018}

\end{document}